# Counterexample-guided Planning

**Krishnendu Chatterjee**
UC Berkeley

**Thomas A. Henzinger**
EPFL

**Ranjit Jhala**
UC San Diego

**Rupak Majumdar**
UC Los Angeles

## Abstract

Planning in adversarial and uncertain environments can be modeled as the problem of devising strategies in stochastic perfect information games. These games are generalizations of Markov decision processes (MDPs): there are two (adversarial) players, and a source of randomness. The main practical obstacle to computing winning strategies in such games is the size of the state space. In practice therefore, one typically works with *abstractions* of the model. The difficulty is to come up with an abstraction that is neither too coarse to remove all winning strategies (plans), nor too fine to be intractable. In verification, the paradigm of *counterexample-guided abstraction refinement* has been successful to construct useful but parsimonious abstractions *automatically*. We extend this paradigm to *probabilistic* models (namely, perfect information games and, as a special case, MDPs). This allows us to apply the counterexample-guided abstraction paradigm to the AI planning problem. As special cases, we get planning algorithms for MDPs and deterministic systems that automatically construct system abstractions.

## 1 Introduction

Sequential decision making under uncertainty is a central problem in artificial intelligence and operations research. The problem involves designing sequences of actions ("plans") in order to achieve certain goals. Markov decision processes (MDPs) [18] have been extensively used to model such decision problems in the presence of uncertainty [3, 13], and have been augmented to the more general framework of two player perfect information stochastic games [8, 9]. A perfect information game consists of a directed graph with three kinds of nodes: player 1 nodes, player 2 nodes, and random choice nodes. At a player 1 (resp. 2) node, player 1 (resp. 2) decides to take an edge to a neighboring node; at a random node, the game proceeds to the neighboring nodes according to a given probability distribution. A *plan* is now a sequence of actions of player 1 such that, no matter how player 2 behaves, player 1 can achieve a certain goal. This model generalizes MDPs by separating adversarial (player 2) from random (player random) choice of the opponent or environment. The model is useful to represent *robust* stochastic optimization and control problems [9]: player 2 models worst-case uncertainty in the transition matrices and rewards, hence a player 1 plan to achieve a goal irrespective of player 2 actions is robust with respect to modeling defects.

While adopted as a useful and robust model, computational issues remain the main hurdle in solving real-life planning problems. This is especially so in situations where systems are compactly represented and state spaces grow exponentially in the number of modeled features (the "curse of dimensionality"). The key to the success of algorithmic planning under uncertainty is *abstraction*. An abstract state space clusters concrete states of the problem, ignoring irrelevant features, and summarizing information about other features. Useful abstractions have two desirable properties. First, the abstraction should be *sound*, meaning that if a property (e.g., existence of a plan with a certain reward) is proved for the abstract model of a system, then the property holds also for the concrete system. Second, the abstraction should be *effective*, meaning that the abstract model is not too fine and can be handled by the tools at hand. Indeed, abstraction has been successfully applied in decision-theoretic planning [2, 14, 7]. In order to be practical, abstractions must have a third desirable property. A sound and effective abstraction (provided it exists) should be found *automatically*; otherwise, the labor-intensive process of constructing suitable abstract models often negates the benefits of automatic methods. Research has focused on the automatic extraction of abstractions: the main technique is to identify a subclass of *relevant* features, and perform a cone-of-influence computation to add other features of the problem as relevant if they directly affect some features that are already known to be relevant. Once a set of relevant features has been thus computed, the abstraction clusters together all states that agree on the valuations of the relevant features. This abstraction (which is usually much smaller than the original problem) is then solved to generate a plan. However, if the abstraction is too coarse, then the generated abstract plan may be far from an optimal solution. In that case, one has to start again with a new set of relevant features and continue.

Instead, we apply a successful paradigm from formal verification to the planning domain: the method of *counterexample-guided abstraction refinement* (CEGAR) [15, 5]. In CEGAR, one starts with a very coarse abstract model, which is effective but may not be informative, meaning that it may produce extremely suboptimal plans. Then the abstract model is refined iteratively as follows: first, if the abstract model does not exhibit the desired property, then an *abstract counterexample* (that exhibits why the property fails on the abstract model) is constructed automatically; second, it can be checked automatically if the abstract counterexample corresponds to a concrete counterexample; if this is not the case, then third, the abstract model is refined automatically in order to eliminate the spurious counterexample. Counterexample-guided abstraction refinement improves on cone-of-influence refinement in that new features are added only at states where they are needed for a plan to achieve the objective, whereas states that need not be distinguished for planning remain lumped together in the abstraction. In this way, a non-uniform abstraction is constructed—exposing more detail where necessary, and less detail where sufficient—and the construction is fully automatic (algorithmic).

The method of counterexample-guided abstraction refinement has been developed for the verification [15, 5] and control [10] of deterministic systems, and applied successfully to software verification [1, 11]. We generalize the technique to work on perfect information games. In verification, a counterexample to the satisfaction of a linear-time property (a temporal property of *traces*) is a possibly infinite trace that violates the property; in control, counterexamples are *strategy trees*. Our first insight is that for probabilistic perfect information games with discounted or average reward winning conditions, winning strategies for both players are pure (deterministic) and memoryless, and hence, counterexamples can be represented as finite objects (graphs). These, in turn, can be iteratively refined.

In somewhat more detail, our method proceeds as follows. Given a perfect information game structure, a winning objective (e.g., discounted reward or average reward), and a goal $p \in \mathbb{R}$, we wish to check if player 1 has a strategy to achieve a value of at least $p$ according to the winning objective, no matter how player 2 behaves. We automatically construct an abstraction of the given game structure that is as coarse as possible and as fine as necessary in order for player 1 to have a strategy that achieves at least $p$ against any strategy of player 2. We start with a very coarse abstract game structure and refine it iteratively. First, we check if player 1 has such a strategy in the abstract game; if so, then she has a strategy in the concrete game; otherwise, we construct an abstract player 2 strategy that spoils against all abstract player 1 strategies (i.e., restricts player 1 to a value less than $p$). Second, we check if the abstract player 2 strategy corresponds to a spoiling strategy for player 2 in the concrete game; if so, then there is no feasible plan for player 1 that achieves value $p$; otherwise, we refine the abstract game in order to eliminate the abstract player 2 strategy. In this way, we automatically synthesize "maximally abstract" plans, which distinguish two states of the system only if they need to be distinguished in order to achieve the winning objective. While we obtain our results on the general framework of perfect information games, as special cases, we get counterexample-guided planning algorithms for planning problems on Markov decision processes, and on deterministic games and transition systems.

## 2 Games and Abstraction

**Perfect information games.** A (*perfect information*) *game structure* $\mathcal{G} = (V, E, v_0, \text{wt}, r, (V_1, V_2, V_r))$ consists of a directed graph $(V, E)$, an initial state $v_0 \in V$, an edge weight function $\text{wt} : E \to (0, 1]$, a reward function $r : V \to \mathbb{R}$, and a partition $(V_1, V_2, V_r)$ of the set of states $V$ into player 1 states $V_1$, player 2 states $V_2$, and random states $V_r$. In the sequel, $i$ ranges over the set $\{1, 2\}$ of players. We require that for every $v \in V_r$, we have $\sum_{v':(v,v')\in E} \text{wt}(v, v') = 1$, that is, the edge weight function determines a probability distribution over states for each random state $v \in V_r$. Weight functions are defined for all edges for notational convenience, we shall not use weight functions on edges outgoing from a player $i$ state. For technical convenience, we require that all states have at least one outgoing edge. Intuitively, at state $v \in V_i$, player $i$ chooses an outgoing edge $(v, v')$ and the game proceeds to $v'$. At a random state $v \in V_r$, the game proceeds to a neighbor $v' \in V$ with probability $\text{wt}(v, v')$. Perfect information game structures subsume several important special cases. If $V_2 = \emptyset$, we have a *Markov decision process* [18]. If $V_r = \emptyset$, we have a *turn-based deterministic game* [20]. If both $V_2$ and $V_r$ are $\emptyset$, we have a transition system.

A *run* of the game structure $\mathcal{G}$ is an infinite sequence $v_0 v_1 v_2 \ldots$ of states $v_j \in V$ such that for all $j \geq 0$, we have $(v_j, v_{j+1}) \in E$. A *strategy of player $i$* is a partial function $f_i : V^* \cdot V_i \to V$ such that for every state sequence $u \in V^*$ and every state $v \in V_i$, we have $(v, f_i(u \cdot v)) \in E$. Intuitively, a player-$i$ strategy suggests a move for player $i$ given a sequence of states that end in a player-$i$ state. Let $F_i$ denote the set of player $i$ strategies. A player $i$ strategy $f_i$ is *memoryless* if it depends only on the current state, that is, for all $u, u' \in V^*$ and $v \in V_i$, we have $f_i(u \cdot v) = f_i(u' \cdot v)$. A memoryless strategy $f_i$ can be represented as a function $f_i : V_i \to V$.

Given two strategies $f_1$ and $f_2$ of players 1 and 2, the *possible outcomes* $\Omega_{f_1,f_2}(v)$ from a state $v \in V$ are runs: a run $v_0 v_1 v_2 \ldots$ belongs to $\Omega_{f_1,f_2}(v)$ iff $v = v_0$ and for all $j \geq 0$, $v_j \in V_i$ and $v_{j+1} = f_i(v_0 \ldots v_j)$, or $v_j \in V_r$ and $(v_j, v_{j+1}) \in E$. An event is a measurable set of runs. Once a starting state $v$ and strategies $f_1$ and $f_2$ for the two players have been chosen, the probabilities of events are uniquely defined.

**Objectives.** A *game* $(\mathcal{G}, \Gamma)$ consists of a game structure $\mathcal{G}$ and an objective $\Gamma$ for player 1. We consider *discounted reward* and *average reward* objectives [18, 8]. Given a strategy $f_1$ for player 1 and $f_2$ for player 2 the values of the game under strategies $f_1$ and $f_2$ from a state $v \in V$ is defined as follows:

- *Discounted reward objective.* Given a discount factor $\beta \in (0,1)$ the values $\mathrm{val}_1^{f_1,f_2}$ and $\mathrm{val}_2^{f_1,f_2}$ for player 1 and player 2 are defined as follows:

$$\mathrm{val}_1^{f_1,f_2}(v) = \sum_{t=0}^{\infty} \beta^t \mathbf{E}_v^{f_1,f_2}[r(v_t)];$$

$$\mathrm{val}_2^{f_1,f_2}(s) = -\sum_{t=0}^{\infty} \beta^t \mathbf{E}_s^{f_1,f_2}[r(v_t)].$$

- *Average reward objective.* The values $\mathrm{val}_1^{f_1,f_2}$ and $\mathrm{val}_2^{f_1,f_2}$ for player 1 and player 2 for average reward objective are defined as follows:

$$\mathrm{val}_1^{f_1,f_2}(v) = \liminf_{N \to \infty} \frac{1}{N} \sum_{t=0}^{N} \mathbf{E}_v^{f_1,f_2}[r(v_t)];$$

$$\mathrm{val}_2^{f_1,f_2}(s) = -\liminf_{N \to \infty} \frac{1}{N} \sum_{t=0}^{N} \mathbf{E}_s^{f_1,f_2}[r(v_t)].$$

**Values.** The values of a game $(\mathcal{G}, \Gamma)$ for player 1 ($\mathrm{val}_1$) and player 2 ($\mathrm{val}_2$) are defined as follows:

$$\begin{aligned} \mathrm{val}_1(v) &= \sup_{f_1 \in F_1} \inf_{f_2 \in F_2} \mathrm{val}_1^{f_1,f_2}(v); \\ \mathrm{val}_2(v) &= \sup_{f_2 \in F_2} \inf_{f_1 \in F_1} \mathrm{val}_2^{f_1,f_2}(v). \end{aligned}$$

A strategy $f_1$ is *optimal for player 1* from a state $v$ if for all strategies $f_2$ of player 2, we have $\mathrm{val}_1^{f_1,f_2}(v) \geq \mathrm{val}_1(v)$. A strategy $f_1$ is *profitable* compared to a strategy $f_1'$ from a state $v$ if we have $\inf_{f_2 \in F_2} \mathrm{val}_1^{f_1,f_2}(v) > \inf_{f_2 \in F_2} \mathrm{val}_1^{f_1',f_2}(v)$. For a real $p \in \mathbb{R}$, a strategy $f_1$ is *$p$-optimal* from state $v$ if for all strategies $f_2$ of player 2 we have $\mathrm{val}_1^{f_1,f_2}(v) \geq p$. A strategy is $p$-optimal if it is $p$-optimal from $v_0$. The following result is classical [19, 16, 8].

**Proposition 1 (Determinacy and optimal strategies).** *Let $\mathcal{G} = ((V,E), v_0, \mathrm{wt}, r, (V_1, V_2, V_r))$ be a game structure, and $\Gamma$ a discounted reward or average reward objective. Then the following assertions hold.*

1. *For all states $v \in V$, we have $\mathrm{val}_1(v) + \mathrm{val}_2(v) = 0$.*

2. *Memoryless optimal strategies exist for player 1 and player 2 from every state $v \in V$.*

The existence of non-randomized memoryless optimal strategies in perfect information games can be contrasted with concurrent stochastic games (games where players make moves simultaneously at each state) where optimal strategies may require both randomization and memory. A function $f_1 : V_1 \to V$ is also called a *(memoryless) plan*. A plan is *$p$-feasible* if (1) it is a strategy for player 1, and (2) it is $p$-optimal. The *planning problem* takes as input a game $(\mathcal{G}, \Gamma)$ and a value $p \in \mathbb{R}$, and produces a $p$-feasible plan for player 1 in the game $(\mathcal{G}, \Gamma)$, or states that no such plan exists. Observe that if a memoryless strategy $f_1$ is fixed for player 1 then a perfect information game reduces to a MDP. The values of a MDP with discounted reward and average reward objectives can be computed in polynomial time (details in [8, 18]). A memoryless optimal strategy represents a polynomial witness and the polynomial time algorithms for MDPs is the polynomial time verification procedure. This establishes that the problem is in NP. Moreover, since the problem is symmetric for both players it follows that the problem is also in co-NP. This gives us the following result.

**Proposition 2 (Complexity).** *Let $\mathcal{G}$ be a perfect information game structure, $\Gamma$ a discounted reward or average reward objective, and $p \in \mathbb{R}$. The planning problem $(\mathcal{G}, \Gamma, p)$ can be solved in exponential time. The complexity of the planning problem is NP $\cap$ co-NP.*

**Abstractions of games.** Since solving a game may be expensive, we wish to construct sound abstractions of the game with smaller state spaces. Soundness means that if player 1 has a $p$-optimal strategy in the abstract game, then she also has a $p$-optimal strategy in the original, concrete game. To ensure soundness, we restrict the power of player 1 and increase the power of player 2 [12, 10]. Therefore, we abstract the player 1 states so that *fewer* moves are available, and the player 2 states so that *more* moves are available. Informally, abstraction represents imprecise information about some (possibly irrelevant) state variables. We do not abstract uncertainties in transitions. Hence, random states (that represent transition uncertainties) are not abstracted.

An *abstraction* $\mathcal{G}^\alpha$ for the game structure $\mathcal{G}$ is a game structure $((V^\alpha, E^\alpha), v_0^\alpha, \mathrm{wt}^\alpha, r^\alpha, (V_1^\alpha, V_2^\alpha, V_r^\alpha))$ and a concretization function $\llbracket \cdot \rrbracket \colon V^\alpha \to 2^V$ such that conditions (1)–(5) hold.

1. The abstraction preserves the player structure: for $i \in \{1, 2\}$ and all $v^\alpha \in V_i^\alpha$, we have $\llbracket v^\alpha \rrbracket \subseteq V_i$; also, random states are not abstracted, that is, $V_r^\alpha = V_r$ and $\llbracket v \rrbracket = \{v\}$ for all $v \in V_r^\alpha$.

2. The abstract states partition the concrete state space: $\bigcup_{v^\alpha \in V^\alpha} \llbracket v^\alpha \rrbracket = V$. Moreover, $v_0^\alpha$ is the (unique) abstract state such that $v_0 \in \llbracket v_0^\alpha \rrbracket$.

3. For each player 1 abstract state $v^\alpha \in V_1^\alpha$, define $(v^\alpha, w^\alpha) \in E^\alpha$ iff for all $v \in \llbracket v^\alpha \rrbracket$ there is a $w \in \llbracket w^\alpha \rrbracket$ such that $(v, w) \in E$. For each abstract state $v^\alpha \in V_2^\alpha \cup V_r^\alpha$, define $(v^\alpha, w^\alpha) \in E^\alpha$ iff there exists $v \in \llbracket v^\alpha \rrbracket$ and $w \in \llbracket w^\alpha \rrbracket$ such that $(v, w) \in E$.

4. The abstraction preserves the probability distribution from random states. For each edge $(v^\alpha, w^\alpha) \in E^\alpha$, define $\mathrm{wt}(v^\alpha, w^\alpha) = \sum \{\mathrm{wt}(v, w) \colon v \in \llbracket v^\alpha \rrbracket, w \in \llbracket w^\alpha \rrbracket\}$.

5. For each abstract state $v^\alpha \in V^\alpha$, define $r^\alpha(v^\alpha) = \min\{r(v) \colon v \in \llbracket v^\alpha \rrbracket\}$.

Note that the abstract state space $V^\alpha$ and the concretization function $\llbracket \cdot \rrbracket$ uniquely determine the abstraction $\mathcal{G}^\alpha$. Intuitively, each abstract state $v^\alpha \in V^\alpha$ represents a set $\llbracket v^\alpha \rrbracket \subseteq V$ of concrete states.

**Proposition 3 (Soundness of abstraction).** *Let $\mathcal{G}^\alpha$ be an abstraction for a game structure $\mathcal{G}$, and let $\Gamma$ be an objective for player 1. For every $p \in \mathbb{R}$, if player 1 has a $p$-optimal strategy in the abstract game $(\mathcal{G}^\alpha, \Gamma)$, then player 1 also has a $p$-optimal strategy in the concrete game $(\mathcal{G}, \Gamma)$.*

The above proposition follows from the fact that abstractions preserve strategies for the players in the following sense: if $f_1^\alpha$ is a strategy for player 1 in an abstraction, then there is a corresponding strategy $f_1$ for player 1 in the concrete game; and if there is a strategy $f_2$ for player 2 in the concrete game, then there is a corresponding strategy $f_2^\alpha$ for player 2 in the abstract game. Hence in the abstraction player 1 is *weakened* and player 2 is *strengthened*. It follows that if there is a $p$-optimal strategy for player 1 in the abstract game, then there is a $p$-optimal strategy for player 1 in the concrete game. This establishes the soundness of the abstraction.

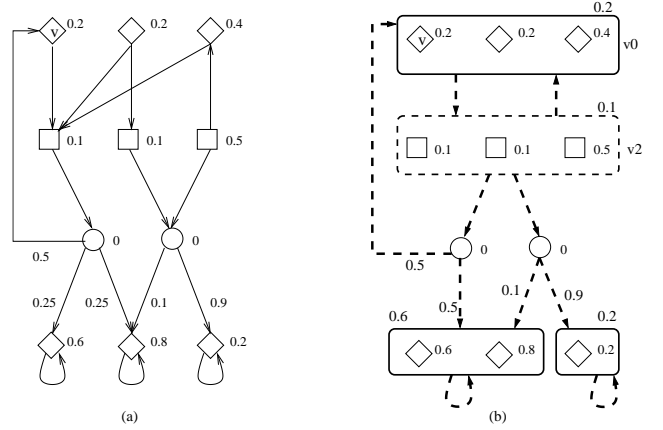

Figure 1: (a) A perfect information game and (b) its abstraction.

**Example 1** Consider the perfect information game structure and its abstraction shown in Figure 1(a). There are eleven states. The diamond states are player 1 states, the box states are player 2 states, the circle states are random states. The state marked $v$ is the initial state. Each state is marked with a reward. We only write the edge weights for edges coming out of random states. Notice that player 1 has a strategy to get 0.7 in the average reward game from state $v$. Figure 1(b) shows one particular abstraction for this game structure. The boxes denote abstract states, with the concrete states they represent drawn inside them. Solid (resp. dashed) boxes denote player 1 (resp. player 2) states. The dashed arrows are the abstract transitions. In the abstract game, the player 2 memoryless strategy that chooses the $v_2 \to v_0$ at state $v_2$ ensures player 1 can only achieve an average reward of 0.15. ∎

## 3 Counterexample Guided Planning

In general, an abstraction loses information, and are not complete. That is, since we provide more power to player 2, it may well be that player 1 has no $p$-optimal strategy in the abstract game, but has a $p$-optimal strategy in the concrete game. We now show how in this case, a *counterexample* can be used to refine the abstraction.

We first informally present the idea of *counter-example guided abstraction-refinement*. A *counterexample* to the claim that player 1 can achieve a reward $p$ is a *spoiling* strategy for player 2 that prevents player 1 from getting $p$. It follows from Proposition 1 that if there is a spoiling strategy then there is a memoryless spoiling strategy. A counterexample for an abstract game $(\mathcal{G}^\alpha, \Gamma)$ may be either *genuine*, meaning that it corresponds to a player 2 strategy for the concrete game $(\mathcal{G}, \Gamma)$, or *spurious*, meaning that it arises due to *coarseness* in abstraction. If a counterexample is spurious then we refine the abstraction and the refinement is guided by the counterexample. In the sequel, we denote by $f^\alpha$ a strategy in the abstract game $\mathcal{G}^\alpha$ and by $\mathrm{val}_1^\alpha$ the value function for player 1 in the abstract game $\mathcal{G}^\alpha$.

**Counterexample analysis and refinement.** We now elaborate the procedure CEGAR shown in Figure 1. The procedure CEGAR takes an abstract game $\mathcal{G}^\alpha$, a spoiling strategy $f_2^\alpha$ for player 2 in the abstract game, and a value function $\mathrm{val}_1^\alpha$ for player 1 in the abstract game. It analyzes whether the spoiling strategy $f_2^\alpha$ is spurious (not feasible in the concrete game) and if so it refines the overapproximation in the present abstraction by applying the operators *Focus* and *ValueFocus* according to some strategy (left unspecified). The two operators *Focus* and *ValueFocus* break up abstract states. The operator *Focus* refines the approximation made in the edge relation caused by the abstraction, while *ValueFocus* refines the approximation in the reward function $r$ made by collapsing states.

The operator *Focus* achieves the following:

- for a player 2 state $v^\alpha$ in $\mathcal{G}^\alpha$ it determines which states in $v^\alpha$ can be a part of the spoiling strategy of player 2 in the concrete game. That is, it returns the subset of states of $[\![v^\alpha]\!]$ that have an edge into some state in $[\![f_2^\alpha(v^\alpha)]\!]$. Formally, for $v^\alpha \in V_2^\alpha$, we define $Focus(v^\alpha, f_2^\alpha, \mathrm{val}_1^\alpha) := \{v \in [\![v^\alpha]\!] : \exists w \in V.\ (v, w) \in E \text{ and } w \in [\![f_2^\alpha(v^\alpha)]\!]\}$.

- for a player 1 state $v^\alpha$ in $\mathcal{G}^\alpha$ it determines which of the states in the concrete game has a strategy that is profitable compared to the optimal strategy $f_1^\alpha$ in $\mathcal{G}^\alpha$. This is achieved by checking locally whether there is a state $v \in v^\alpha$ that has a successor in the $\mathcal{G}^\alpha$ with a greater value than $\mathrm{val}_1^\alpha(v^\alpha)$. Formally, for $v^\alpha \in V_1^\alpha$, we define $Focus(v^\alpha, f_2^\alpha, \mathrm{val}_1^\alpha) := \{u \in [\![v^\alpha]\!] : \exists w \in V, w^\alpha \in V^\alpha.(u, w) \in E \wedge w \in [\![w^\alpha]\!] \wedge \mathrm{val}_1^\alpha(w^\alpha) > \mathrm{val}_1^\alpha(v^\alpha)\}$. Note that this step is similar to the policy iteration based on values.

The *ValueFocus* operator is used to split abstract states in $v^\alpha$ in two classes: the set of states $v \in v^\alpha$ such that $r(v) = r^\alpha(v^\alpha)$ and the set of states $v \in v^\alpha$ such that $r(v) > r^\alpha(v^\alpha)$. Formally, $ValueFocus(v^\alpha) := \{v \in v^\alpha\ :\ r(v) = r^\alpha(v^\alpha)\}$.

**Counterexample guided plan generation.** Given a game structure $\mathcal{G}$, an objective $\Gamma$, and a goal $p \in \mathbb{R}$, we wish to determine if the planning problem $(\mathcal{G}, \Gamma, p)$ has a solution, and if so, construct a plan ($p$-optimal strategy) for player 1 ("synthesize a plan"). Our algorithm CounterExampleGuidedPlan, which generalizes the "abstract-verify-refine" loop of [5, 10], proceeds as follows:

**Step 1** ("Abstraction") We first construct an initial abstract game $(\mathcal{G}^\alpha, \Gamma)$.

**Step 2** ("Strategy synthesis") We solve the abstract game to find if player 1 has a $p$-optimal strategy. If so, we get a $p$-optimal player-1 strategy for the abstract game, from which a $p$-optimal player-1 strategy in the concrete game can be constructed. If not, we get an *abstract counterexample* (AC) encoding a memoryless spoiling strategy for player 2. Notice that solving the abstract game can be done using the usual value or policy iteration methods on the abstract game.

**Step 3** ("Counterexample guided abstraction refinement") If solving the abstract game returns an AC $f^\alpha$, then we use the procedure CEGAR($\mathcal{G}^\alpha, f^\alpha, \mathrm{val}_1^\alpha$) to check if the spoiling strategy represented by the AC $f^\alpha$ is feasible (i.e., genuine). If so, then player 2 has a spoiling strategy in the concrete game, and there is no $p$-optimal plan for player 1. If the AC is spurious, then we use the procedure CEGAR($\mathcal{G}^\alpha, f^\alpha, \mathrm{val}_1^\alpha$) to refine the abstraction $\mathcal{G}^\alpha$, so that $f^\alpha$ (and similar counterexamples) cannot arise on subsequent invocations of the game solver.

**Goto Step 2.** The process is iterated until we find either a player-1 $p$-optimal strategy in step 2, or a genuine counterexample in step 3.

The procedure is summarized in Algorithm 2. The function $InitialAbstraction(\mathcal{G}, \Gamma)$ returns a trivial abstraction for $\mathcal{G}$ that merges all player 1 states and all player 2 states, but preserves initial states. The function $GameSolve(\mathcal{G}^\alpha, \Gamma, p)$ returns a pair $(1, f^\alpha, \mathrm{val}_1^\alpha)$ if player 1 has a $p$-optimal strategy from the initial state in the abstract game, where $f^\alpha$ is a (memoryless) $p$-optimal strategy for player 1, and otherwise it returns $(2, f^\alpha, \mathrm{val}_1^\alpha)$, where $f^\alpha$ is an AC.

From the soundness of abstraction (Proposition 3), the counter-example guided abstraction refinement

**Algorithm 1** Algorithm CEGAR

**Input:** an abstraction $\mathcal{G}^\alpha$, a memoryless spoiling strategy $f_2^\alpha$, the value function $\text{val}_1^\alpha$.
**Output:** if $f_2^\alpha$ is spurious, then SPURIOUS and a refined abstraction; otherwise GENUINE.
$R := \{[\![v^\alpha]\!] \ : \ v^\alpha \in V^\alpha\}$.
**choose nondeterministically**
   **if** there is some node $v^\alpha$ such that $v^\alpha \neq \textit{Focus}(v^\alpha)$ **then**
      $R := (R \setminus [\![v^\alpha]\!]) \cup \{([\![v^\alpha]\!] \setminus \textit{Focus}(v^\alpha, f_2^\alpha, \text{val}_1^\alpha)), \textit{Focus}(v^\alpha, f_2^\alpha, \text{val}_1^\alpha)\}$.
      **return** (SPURIOUS, ABSTRACTION (R)).
**or**
   **if** there is some node $v^\alpha$ such that $v^\alpha \neq \textit{ValueFocus}(v^\alpha)$ **then**
      $R := (R \setminus [\![v^\alpha]\!]) \cup \{([\![v^\alpha]\!] \setminus \textit{ValueFocus}(v^\alpha)), \textit{ValueFocus}(v^\alpha)\}$.
      **return** (SPURIOUS, ABSTRACTION (R)).
**return** GENUINE.

---

**Algorithm 2** CounterExampleGuidedPlan($\mathcal{G}, \Gamma, \mathsf{p}$)

**Input:** a game structure $\mathcal{G}$, an objective $\Gamma$, a real $p$.
**Output:** either FEASIBLE and a feasible plan for player 1,
        or INFEASIBLE and a player 2 spoiling strategy.
$\mathcal{G}^\alpha := \textit{InitialAbstraction}(\mathcal{G}, \Gamma)$
**repeat**
   $(winner, f^\alpha, \text{val}_1^\alpha) := \textit{GameSolve}(\mathcal{G}^\alpha, \Gamma, p)$
   **if** $winner = 2$ and $\textsf{CEGAR}(\mathcal{G}^\alpha, \mathsf{f}^\alpha) = (\textsc{Spurious}, \mathcal{H}^\alpha)$
      $\mathcal{G}^\alpha := \mathcal{H}^\alpha; \ winner := \bot$ **endif**
**until** $winner \neq \bot$
**if** $winner = 1$
   **then return** (FEASIBLE, $f^\alpha$)
**return** (INFEASIBLE, $f^\alpha$)

---

and the correctness of policy iteration algorithm to obtain optimal strategies, we get the correctness of the algorithm. For finite-state games, the procedure CounterExampleGuidedPlan terminates, since every refinement step breaks at least one abstract state. In the worst case by successive refinement the algorithm may end up with the concrete game. This establishes the following result.

**Proposition 4 (Partial correctness).** *If the procedure* CounterExampleGuidedPlan($\mathcal{G}, \Gamma, \mathsf{p}$) *returns* (FEASIBLE,$f$), *then player 1 has a p-optimal strategy in the game* $(\mathcal{G}, \Gamma)$ *and $f$ is a feasible plan. If the procedure returns* INFEASIBLE*, then player 1 does not have a p-optimal strategy in* $(\mathcal{G}, \Gamma)$.

**Example 2** Suppose that the planning procedure is called with the input $(\mathcal{G}, \textit{Average}, 0.5)$, where $\mathcal{G}$ is the game structure of Figure 1(a). That is, we wish to find a plan that ensures player 1 gets at least 0.5 reward in the average reward game. Notice that with the initial abstraction of Figure 1(b), the memoryless spoiling strategy of player 2 chooses at state $v_2$ the transition $v_2 \to v_0$ ensures that in the initial abstraction player 1 only receives an average reward of 0.15. So we check if this counterexample is spurious, and refine the abstraction accordingly. Figure 2(a) shows the result of a value focus on the abstract game of Figure 1(b). The state $v_0$ of the initial abstraction is refined to state $v_0$ and $v_1$. Notice that this value focus rules out the current counterexample. However, consider the following memoryless spoiling strategy of player 2 in this new game: choose at state $v_2$ the transition $v_2 \to v_1$. This gives player 1 an average reward of 0.25. When we apply focus to this counterexample, we obtain the new abstraction in Figure 2(b). The state $v_2$ is split into $v_2$ and $v_3$ as the counterexample strategy for player 2 was spurious. On this abstraction, the player 1 strategy that chooses $v_0 \to v_3$ ensures an average reward of 0.6. Hence there is a 0.5-optimal strategy for player 1 in the present abstraction and we conclude that player 1 has a 0.5-optimal plan in the original game. The plan in the abstract game can be used to synthesize a plan in the concrete game. ∎

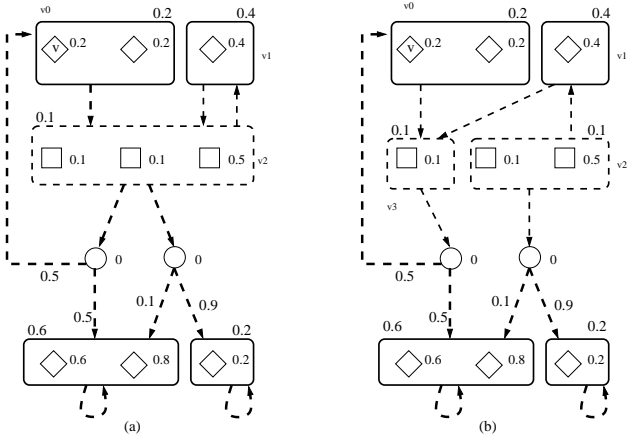

Figure 2: Two steps in the refinement process.

As special cases, we get a counterexample guided plan synthesis algorithm for MDPs ($V_2 = \emptyset$), deterministic games ($V_r = \emptyset$), and transition systems ($V_2 = \emptyset$ and $V_r = \emptyset$).

**Compact representation.** In practice, systems are represented using a compact feature-based representation, and an explicit graph representation (as we have assumed in our exposition) is not available (there is usually an exponential cost in transforming the feature-based representation to the explicit representation). We show how our algorithm can be adapted to a feature-based representation for the important special case of deterministic transition systems.

A transition system is compactly represented by a logical propositional language as follows. Let $\mathcal{P}$ be a set of atomic propositions. A *state* is a valuation to all propositions in $\mathcal{P}$, an *action* is a binary relation on states; these correspond to the vertices and edges of our explicit representation. For a set of propositions $\mathcal{P}$, let $S(\mathcal{P})$ denote the set of valuations over $\mathcal{P}$. Sets of states are compactly represented using a propositional formula over $\mathcal{P}$, and actions are compactly represented as a propositional formula over $\mathcal{P} \cup \mathcal{P}'$, where $\mathcal{P}'$ consists of primed versions of each proposition in $\mathcal{P}$. A *boolean system* $(\mathcal{P}, \mathcal{A})$ consists of a set of propositions $\mathcal{P}$ and a set of compactly represented actions. Similar representations for stochastic actions are possible [7, 6].

The boolean system planning problem asks, given a boolean system $\mathcal{G} = (\mathcal{P}, \mathcal{A})$, an initial state $\varphi_i$ and a final set of states $\varphi_f$ (both represented as propositional formulas over $\mathcal{P}$), whether there exists a sequence of actions in $\mathcal{A}$ that takes the state $\varphi_i$ to some state in $\varphi_f$ (a "feasible plan"), and to construct a feasible plan in this case. We describe the counterexample-guided planning algorithm for the boolean system planning problem. Let $\Pi \subseteq \mathcal{P}$. The set $\Pi$ induces an abstract boolean system $(\Pi, \mathcal{A}[\Pi])$ of a boolean system $(\mathcal{P}, \mathcal{A})$ as follows. An abstract state $s'$ over $\Pi$ is a valuation to propositions in $\Pi$. The concretization $[\![s']\!]$ of an abstract state $s' \in S(\Pi)$ is the set of states $s \in S(\mathcal{P})$ that agree with $s'$ on all propositions in $\Pi$. The abstraction $a[\Pi]$ of an action $a \in \mathcal{A}$ over $\Pi$ is the projection of the action $a$ on to propositions in $\Pi$, i.e., $(s'_1, s'_2) \in a_\Pi$ iff there exist $s_1 \in [\![s'_1]\!]$ and $s_2 \in [\![s'_2]\!]$ such that $(s_1, s_2) \in a$. Let $\mathcal{A}[\Pi] = \{a[\Pi] : a \in \mathcal{A}\}$. For a state $s \in S(\mathcal{P})$, let the abstraction $s[\Pi]$ be the projection of $s$ on to propositions in $\Pi$. The abstraction of a set $S' \subseteq S(\mathcal{P})$ is the set $S'[\Pi] = \{s[\Pi] : s \in S'\}$. The abstraction of a boolean system can be effectively computed from the description of the boolean system by existential quantification. Notice that an abstract boolean system has more behaviors than the concrete boolean system. Thus, if there is no feasible plan that takes $\varphi_i[\Pi]$ to some state in $\varphi_f[\Pi]$ in the abstraction, then there is no feasible plan that takes $\varphi_i$ to some state in $\varphi_f$ in the concrete system. On the other hand, an abstractly feasible path may not be concretely feasible. The counterexample guided planning algorithm, a variation of the algorithm in [5], proceeds as follows.

**Step 1** ("Abstraction") At each step, we maintain an abstract boolean system $(\Pi, \mathcal{A}[\Pi])$ induced by a set of abstraction variables $\Pi \subseteq \mathcal{P}$. The initial abstraction is $(\emptyset, \mathcal{A}[\emptyset])$.

**Step 2** ("Abstract plan synthesis") We solve a reachability problem in the abstract boolean system to find if there is a sequence of abstract actions $a_1[\Pi]a_2[\Pi]\ldots a_n[\Pi]$ that takes $\varphi_i[\Pi]$ to some state in $\varphi_f[\Pi]$. If there is no such sequence, then we stop and return INFEASIBLE. We can implement this reachability algorithm using symbolic data structures such as BDDs [4].

**Step 3** ("Counterexample refinement") Given a sequence of abstract actions $a_1[\Pi]a_2[\Pi]\ldots a_n[\Pi]$, we check if this sequence corresponds to a feasible plan in the concrete system. This check is reduced to a boolean satisfiability check as follows. First, we construct the corresponding sequence of concrete actions $a_1 a_2 \ldots a_n$. Recall that each action $a_i$ is a propositional formula over the set of propositions $\mathcal{P} \cup \mathcal{P}'$, let us make this explicit by writing $a_i(\mathbf{x}, \mathbf{x}')$. We construct the boolean formula

$$\varphi_i(\mathbf{x_1}) \wedge \bigwedge_{i=1}^{n} \mathbf{a_i}(\mathbf{x_i}, \mathbf{x_{i+1}}) \wedge \varphi_f(\mathbf{x_{n+1}})$$

which is satisfiable iff the plan $a_1 \ldots a_n$ is feasible. Notice that we rename the propositional variables along the path. We check the satisfiability of this formula using an efficient boolean satisfiability procedure. If the formula is satisfiable, we return FEASIBLE, and the feasible plan $a_1 \ldots a_n$. If the formula is not satisfiable, the SAT solver constructs a resolution proof that shows why the formula is not satisfiable [21]. The refined abstraction is obtained by adding all variables appearing in this resolution proof to the set of abstraction variables [5, 17]. This ensures that the current infeasible plan is ruled out in subsequent iterations.

**Iterate** The three steps are iterated with the refined abstraction until we find either a feasible plan (in Step 3), or show that no feasible plan is possible (in Step 2). Since at least one variable is added in each iteration, the iteration is bound to terminate.

With suitable algorithms for reachability (Step 2) and satisfiability (Step 3), the same algorithm generalizes to MDPs and games. Moreover, using data structures for compact representations for MDPs [6], the algorithm can be made completely symbolic.

**Acknowledgments.** This research was supported in part by the ONR grant N00014-02-1-0671, the AFOSR MURI grant F49620-00-1-0327, and the NSF grants CCR-0225610, CCR-0234690, and CCR-0427202.